\documentclass[conference]{IEEEtran}
\IEEEoverridecommandlockouts
\usepackage{cite}
\usepackage{amsmath,amssymb,amsfonts}
\usepackage{algorithmic}
\usepackage{graphicx}
\usepackage{textcomp}
\usepackage{ifpdf}
\usepackage{xcolor}
\usepackage[printonlyused,nolist]{acronym}
\usepackage{subfigure}
\usepackage{svg}
\usepackage{comment}
\usepackage{amsmath}
\def\BibTeX{{\rm B\kern-.05em{\sc i\kern-.025em b}\kern-.08em
    T\kern-.1667em\lower.7ex\hbox{E}\kern-.125emX}}

\begin{document}

% Define each acronym individually without creating a list
\newacro{DRL}{Deep Reinforcement Learning}
\newacro{RL}{Reinforcement Learning}
\newacro{RRT}{Rapidly-exploring Random Trees}
\newacro{BFS}{Breadth First Search}
\newacro{FAO}{World Food and Agriculture Organization}
\newacro{4WIS4WID}{Four-Wheel-Independent Steering and Four-Wheel Independent Driving}
\newacro{4WS}{Four Wheel Steering}
\newacro{ROS}{Robot Operating System}
\newacro{TDP}{Technology Development Program}
\newacro{DST}{Department of Science and Technology}
\newacro{URDF}{Unified Robotics Description Format}
\newacro{PID}{Proportional Integral Derivative}
\newacro{KF}{Kalman Filter}
\newacro{EKF}{Extended Kalman Filter}
\newacro{PF}{Particle Filter}

\title{Autonomous Navigation of 4WIS4WID Agricultural Field Mobile Robot using Deep Reinforcement Learning\\
\author{Tom Baby$^{1}$, Mahendra Kumar Gohil$^{1}$, and Bishakh Bhattacharya$^{1}$% <-this % stops a space
\thanks{$^{1}$Department of Mechanical Engineering, Indian Institute of Technology Kanpur, Kanpur 208016, India.}}}%}
\maketitle
\begin{abstract}
In the futuristic agricultural fields compatible with Agriculture 4.0, robots are envisaged to navigate through crops to perform functions like pesticide spraying and fruit harvesting, which are complex tasks due to factors such as non-geometric internal obstacles, space constraints, and outdoor conditions. In this paper, we attempt to employ Deep Reinforcement Learning (DRL) to solve the problem of 4WIS4WID mobile robot navigation in a structured, automated agricultural field. This paper consists of three sections: parameterization of four-wheel steering configurations, crop row tracking using DRL, and autonomous navigation of 4WIS4WID mobile robot using DRL through multiple crop rows. We show how to parametrize various configurations of four-wheel steering to two variables. This includes symmetric four-wheel steering, zero-turn, and an additional steering configuration that allows the 4WIS4WID mobile robot to move laterally. Using DRL, we also followed an irregularly shaped crop row with symmetric four-wheel steering. In the multiple crop row simulation environment, with the help of waypoints, we effectively performed point-to-point navigation. Finally, a comparative analysis of various DRL algorithms that use continuous actions was carried out.

\end{abstract}


\section{Introduction}

The global population is expanding at a rapid rate along with the demand for sustenance. \acf{FAO} predicts that the global population will reach 9.7 billion by 2050\cite{RePEc:ags:faoeff:319842}. To sustain this population, food production must be considerably enhanced. Traditional farming practices and the ever-decreasing availability of farm labour render manual agriculture economically untenable and inefficient. Also, due to severe resource constraints, there is an immediate need to address issues such as the indiscriminate use of agrochemicals, energy conservation, environmental contamination management, and the effects of global warming.
 
 Improving the quantity and quality of agricultural products necessitates research into the development of intelligent, self-sufficient machinery for agricultural tasks. Field Robot-based automation in agriculture has significant socioeconomic effects on consumers, as it reduces food production costs.
 Agricultural automation can increase productivity, enhance product quality and resource-use efficiency, reduce human drudgery and labour shortages, and improve environmental sustainability. However, manoeuvring agricultural field mobile robots through crop rows and avoiding sensitive obstacles like tender plants under outdoor conditions remains highly challenging today.

 This paper aims to present an efficient navigation strategy for a \acf{4WIS4WID} mobile robot for agricultural fields where crops are planted in an organized manner. The contribution includes parameterization of some of the configurations possible in \acf{4WS}. This work also incorporates implementing deep reinforcement learning as a solution for crop row tracking and navigation across multiple crop rows.

\section{RELATED WORKS}

A 4WIS4WID mobile robot has four steering actuators without mechanical links and four wheels that can operate independently. It is capable of movements comparable to those of omnidirectional mobile robots. In contrast to omnidirectional robots, however, the 4WIS4WID robot requires wheel orientation changes and cannot move in any direction immediately\cite{https://doi.org/10.1049/joe.2014.0241}. Overall, the 4WIS4WID mobile robot incorporates the benefits of various steering configurations, providing enhanced manoeuvrability and the ability to perform a variety of motions. 

Vision-based navigation systems are likely the most cost-effective and reliable source of information when compared to all other sensors developed to date\cite{BAI2023107584}. Hough transform has significant anti-interference ability and robustness and is less affected by noise in agricultural environments \cite{ZINEELABIDINE202130}. However, the application of the conventional Hough transform in real-time systems is severely constrained due to the extensive computation required. The OpenCV Probabilistic Hough Transform implementation\cite{MATAS2000119} demonstrates that the proposed algorithm minimizes the computation required to detect lines by exploiting the difference in the fraction of votes required to detect lines.

Precision agriculture jobs often require point-to-point planning. In a complicated context, coverage path planning is not always viable because it is computationally demanding and is primarily utilized in agricultural field machines\cite{9096177}. In Cell Decomposition navigation with D* Lite, the path cannot be precisely followed by a robot with medium-high computation\cite{8228012}. For large-dimensional terrain, navigation utilizing the A* star algorithm is memory-intensive\cite{santos_santos_mendes_costa_lima_reis_shinde_2020}. Current SLAM methods can only be applied precisely in a simplified environment\cite{8456505}. High computational time is associated with reactive path planning approaches. Existing reactive path planning algorithms have not yet attained the required level of robustness and dependability \cite{PATLE2019582}. The methods as mentioned earlier use a map of the environment to plot a course for the robot to follow as it travels to its destination. However, the 2D mapping methods are only suitable for indoor and flat terrain because it is very challenging to create a precise model of a dynamic external environment\cite{JIANG202017}. 

Traditional motion planning approaches are frequently inadequate for completing jobs in complex outdoor settings. Global path planning for robots with RL shows better performance in terms of path length and path smoothness compared to \acf{RRT} and \acf{BFS} \cite{10.1109/ROBIO49542.2019.8961753}. \acf{DRL} has also been used to handle the challenge of motion planning in a high-dimensional environment, and significant progress has been made in this area. In \ac{DRL}, we don't have to annotate vast amounts of data. \ac{DRL} implementation challenges include creating an effective reward function. However, it still has the problem of poor generalization to previously unseen facts \cite{9419029}, which can be tackled using domain randomization. \ac{DRL} was applied to the problem of robot navigation in environments with unknown, rough terrain and varying environment size and traversability \cite{8468643}. The classical methods had the lowest success rate in this work. The DRL method naturally accommodates robot-terrain interactions during the learning process due to its end-to-end approach\cite{9468918}. Autonomously navigating through crop rows without maintaining any maps using a skid-steered wheeled mobile robot is successfully carried out in \cite{9197114}. However, this navigation scheme implies that the robot's starting position is in one of the field's corners and does not account for point-to-point navigation.

Most current research has concentrated on using easily accessible ROS models for skid-steered or differential drive robots in path-planning investigations. Nevertheless, these studies lack consideration of both kinematic and dynamic constraints. In addition, there is a considerable gap of research on three-dimensional path planning and robots navigating through internal non-geometric obstacles. Furthermore, there is a critical gap in the current research landscape as there are few studies on the 4WIS4WID configuration combined with Deep Reinforcement Learning (DRL).

Even though DRL represents a state-of-the-art technique for autonomous navigation, its use in intricate robot configurations and three-dimensional environments has not received much attention. Given these substantial gaps in the existing literature, the primary purpose of this paper is to delve into these critical research areas. It aims to contribute by exploring and advancing navigation strategies that specifically address kinematic constraints, three-dimensional environments, navigation through internal non-geometric obstacles with the help of a camera, and the integration of DRL with the 4WIS4WID robot configurations. Through this, the paper endeavours to not only fill these critical research gaps but also enhance the state-of-the-art in autonomous navigation for an agricultural field.

\section{METHODOLOGY}

\subsection{Robot Setup}\label{robot_setup}

\begin{figure}[t]
    \centering
    \includegraphics[scale=0.4]{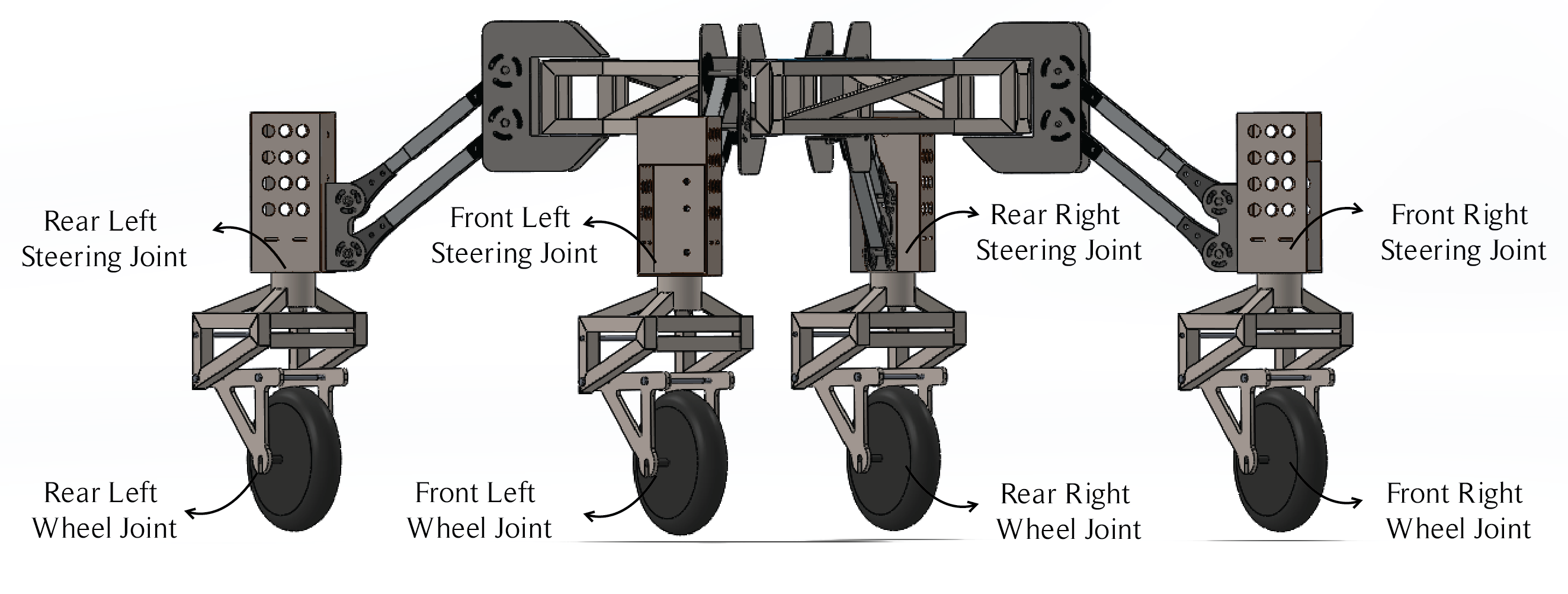}
    \caption{3D model of 4WIS4WID field mobile robot used in simulation.}
    \label{fig:0}
\end{figure}

The 3D model of the robot we have used in the simulation is shown in Figure \ref{fig:0}. There are eight joints, four revolute joints take care of the steering, and four continuous joints are incorporated into the wheels. Steering joints are limited in the range $\left[ -90^{0},90^{0}\right]$.

\begin{figure}[t]
    \centering
    \includegraphics[scale=0.45]{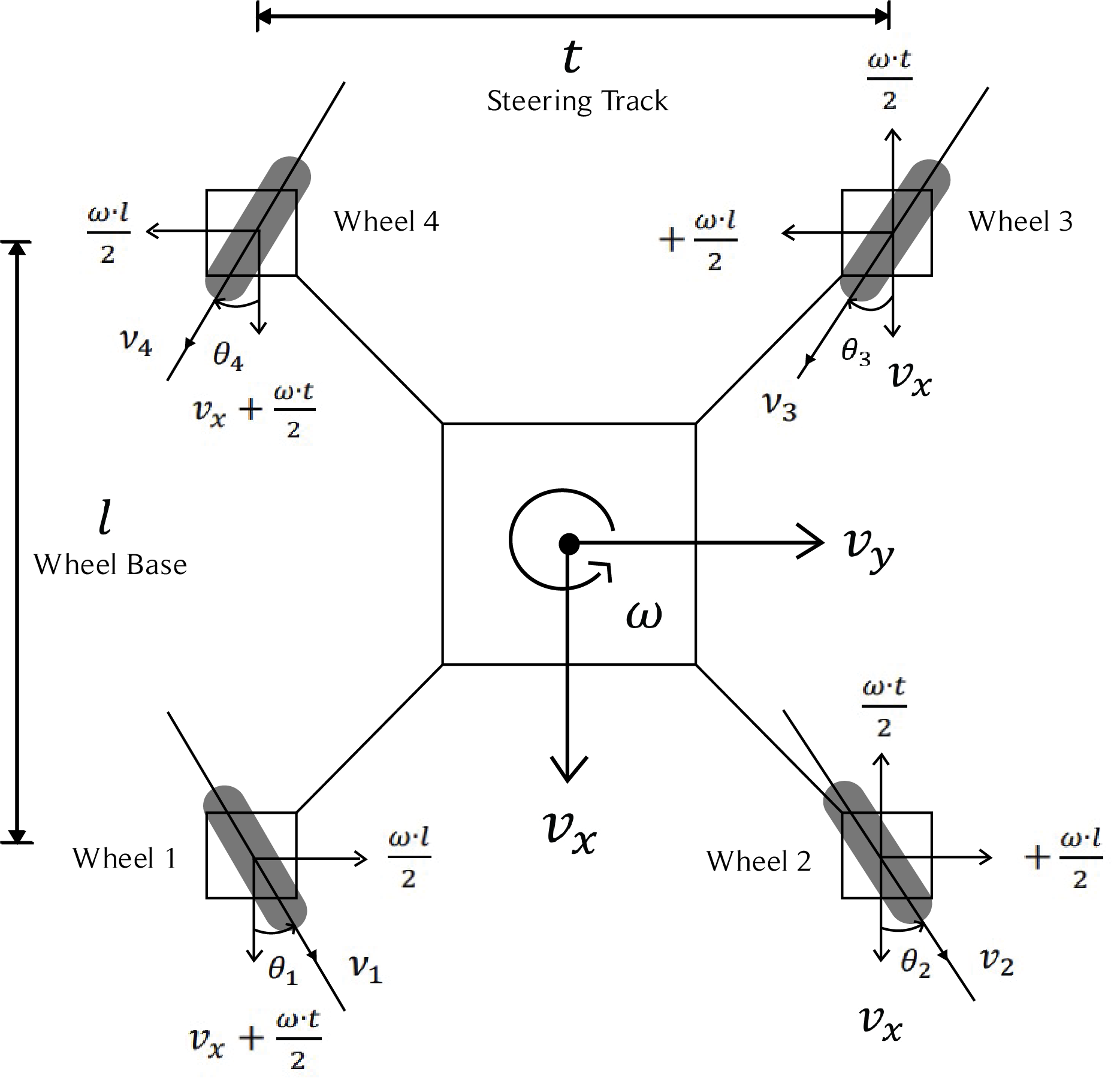}
    \caption{Configuration of Symmetric \ac{4WS}.}
    \label{fig:1}
\end{figure}

We assume that this 4WIS4WID mobile robot is rigid. We account for 3 Degrees of Freedom without contemplating the pitch, roll, and heave motions, as depicted in Figure \ref{fig:1}. The position of the mobile robot's centre of mass and centroid are identical. $v_x$ and $v_y$ are the linear velocities in the global coordinate system, while $\omega$ is the angular velocity. It is assumed that the radius of all the wheels is identical. Since we establish separate steering for lateral movement, we presume no-slip condition when deriving equations for symmetric \ac{4WS} and zero turn steering; hence $v_y$ can also be neglected. 

Consider $t$ is the distance between the front Wheels and $l$ is the distance between the front and back Wheels (Wheel Base). These two values are kept equal in our case. $R$ is the wheel radius.$\,v_{1},\,v_{2},\,v_{3,},\,v_{4}$ are corresponding linear velocities at wheel 1, wheel 2, wheel 3, and wheel 4 respectively. 

Each wheel is located at a distance of  $\left( \dfrac{l}{2},\dfrac{t}{2}\right) $ from the centre of the robot.

Squaring and adding the components of global velocity at the wheel 1, we get

\[v1 = \mathrm{signum}\! \left(v_{x}\right) \sqrt{{(\,v_{x} + \frac{{\omega t}}{2})\,^{2}+(\,\frac{{\omega l}}{2})\,^{2}}} \]

The signum function returns the sign of the value inside the function. Consider  wheel 1 in which angular velocity of the corresponding wheel = $\dfrac{v_{1}}{R}$\\

Similarly, we obtain the angular velocities of all the wheels as,

\begin{equation}\label{eq:3.1}
    \omega_{1} = \frac{\mathrm{signum}\! \left(v_{x}\right) \sqrt{{(\,v_{x} + \frac{{\omega t}}{2})\,^{2}+(\,\frac{{\omega l}}{2})\,^{2}}}}{R}\\
\end{equation}
\begin{equation}\label{eq:3.2}
    \omega_{2} = \frac{\mathrm{signum}\! \left(v_{x}\right) \sqrt{{(\,v_{x} - \frac{{\omega t}}{2})\,^{2}+(\,\frac{{\omega l}}{2})\,^{2}}}}{R}\\
\end{equation}
\begin{equation}\label{eq:3.3}
    \omega_{3} = \frac{\mathrm{signum}\! \left(v_{x}\right) \sqrt{{(\,v_{x} - \frac{{\omega t}}{2})\,^{2}+(\,\frac{{\omega l}}{2})\,^{2}}}}{R}\\
\end{equation}
\begin{equation}\label{eq:3.4}
    \omega_{4} = \frac{\mathrm{signum}\! \left(v_{x}\right) \sqrt{{(\,v_{x} + \frac{{\omega t}}{2})\,^{2}+(\,\frac{{\omega l}}{2})\,^{2}}}}{R}\\
\end{equation}

To calculate the steering angle at each wheel, we calculate the arctangent between the resultant velocity vector and the $x-axis$. From Figure \ref{fig:2} we get $\theta_{1}, \theta_{2}, \theta_{3}$ and $\theta_{4}$ respectively as\\

\begin{equation}\label{eq:3.5}
\theta_{1}\ = \arctan \! \left(\frac{\omega  l}{\omega  t +2 v_{x}}\right)\\
\end{equation}
\begin{equation}\label{eq:3.6}
\theta_{2}\ = \arctan\! \left(\frac{\omega  l}{-\omega  t+2 v_{x}}\right)\\
\end{equation}
\begin{equation}\label{eq:3.7}
\theta_{3}\ = -\arctan \! \left(\frac{\omega  l}{-\omega  t +2 v_{x}}\right)\\
\end{equation}
\begin{equation} \label{eq:3.8}
\theta_{4}\ = -\arctan \! \left(\frac{\omega  l}{\omega  t +2 v_{x}}\right) \\
\end{equation}

As we can find from the equations (\ref{eq:3.5}-\ref{eq:3.8}) the pair of $(\theta_{1},\theta_{4})$ and $(\theta_{2},\theta_{3})$  are equal in magnitude, but with opposite signs. As a result, we obtained a symmetric 4WS configuration with symmetrical angles on each side.

The normal lines to the centre of each wheel plane must connect at a single location for the wheels of a 4WS to be slip-free during a turn. This is also known as the kinematic steering condition. From \cite{Jazar2008}, the kinematic steering condition for symmetric steering is given as
\begin{equation}\label{eq:3.9}
	\cot \delta_o-\cot \delta_i=\frac{w_f}{l}+\frac{w_r}{l}
\end{equation}

where $w_f$ and $w_r$ represent the front and rear track length, $\delta_i$ and $\delta_o$ represent the inner and outer wheel steer angles respectively. In our case, the steering track is equal to the wheelbase; hence, the RHS of equation \ref{eq:3.9} reduces to 2. Substituting equations \ref{eq:3.5} and \ref{eq:3.6} for the values in $\delta_i$ and $\delta_o$ in equation \ref{eq:3.9}, we obtain that LHS is also equal to 2. Hence, our symmetric 4WS  satisfies the kinematic steering condition.

To achieve zero turn steering with our model, we substitute the value of global velocity $v_{x}$ to zero in equations \ref{eq:3.1}-\ref{eq:3.8} and obtain

\begin{equation}\label{eq:3.10}
    \omega_1=\omega_4 =  \omega \frac{\left(\sqrt{l^2+t^2}\right)}{2 R} 
\end{equation}

\begin{equation}\label{eq:3.11}
\omega_2=\omega_3=  - \omega \frac{\left(\sqrt{l^2+t^2}\right)}{2 R}
\end{equation}

\begin{equation}\label{eq:3.12}
    \theta_1=\theta_3=\arctan \left(\frac{l}{t}\right) 
\end{equation}

\begin{equation}\label{eq:3.13}
    \theta_2=\theta_4=- \arctan \left(\frac{l}{t}\right)
\end{equation}

As in this case, the steering track is equal to the wheelbase; hence, equations 12 and \ref{eq:3.13} reduce to  

\begin{equation}\label{eq:3.14}
    \theta_1=\theta_3=45^{\circ} \quad\text{and}\quad \theta_2=\theta_4= -45^{\circ}
\end{equation}

The 4WS system in a mobile robot allows lateral movement by turning all the wheels to a 90-degree angle, resulting in sideways motion. We provide equal angular velocities to all the wheels so that the mobile robot can move laterally to the right or left, depending on the direction of rotation.

\begin{equation}
    \theta_1=\theta_2=\theta_3=\theta_4=90^{\circ}
\end{equation}

\begin{equation}
     \omega_1=\omega_2=\omega_3=\omega_4= \pm \omega 
\end{equation}

With the equations developed in this section, teleoperation for different 4WIS4WID steering configurations using a keyboard or joystick can be accomplished.

%\subsection{Single Crop Row Tracking}
\subsection{Crop Row Detection}\label{crop_row_detection}

\begin{figure*}[ht!]
        \centering
            \subfigure[Location of Camera in the Robot Model.]
            {
                \label{subfig:2_1}
                \includegraphics[scale=0.3]{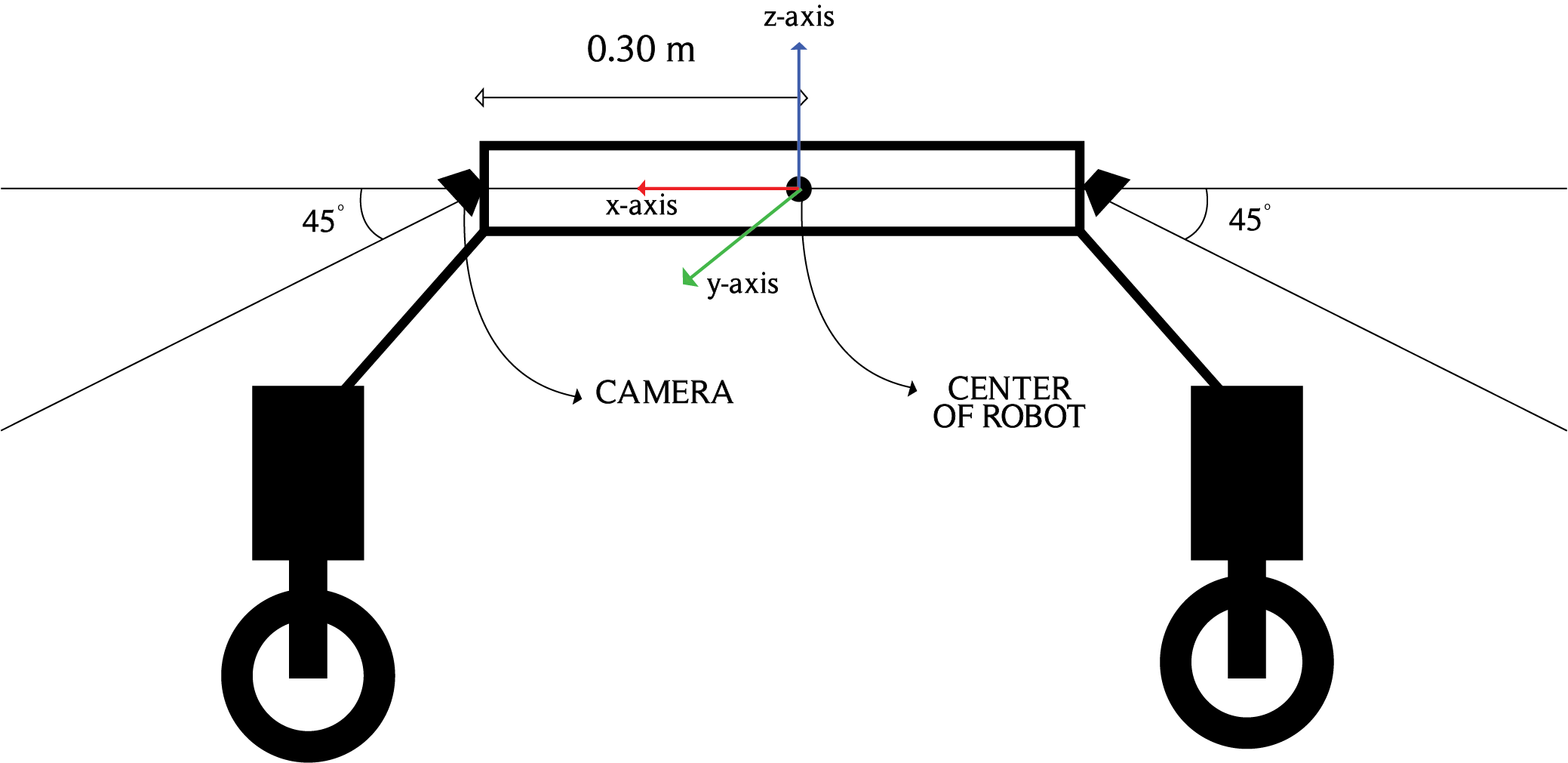} 
                
            } \hspace{1mm}
            \subfigure[Steps in Crop Row Detection.]
            {
                \label{subfig:2_2}
                \includegraphics[scale=0.4]{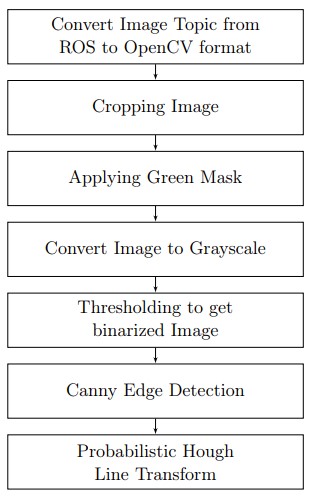}
            }\hspace{1mm}
            \subfigure[Crop row track error.]
            {
                \label{subfig:2_3}
                \includegraphics[scale=0.76]{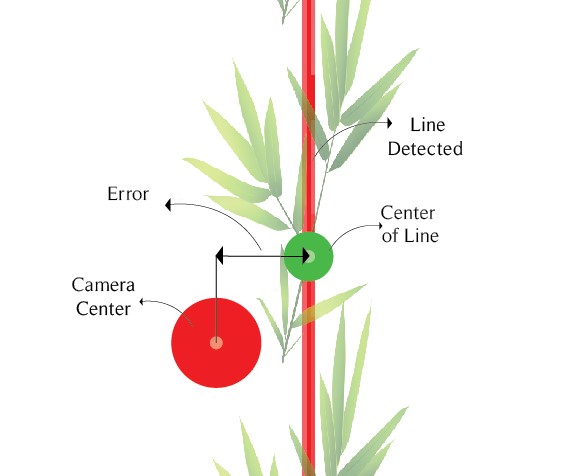}
            }\hspace{1mm}
            \subfigure[Example of Waypoint generation.]
            {
                \label{subfig:2_4}
                \includegraphics[scale=0.2]{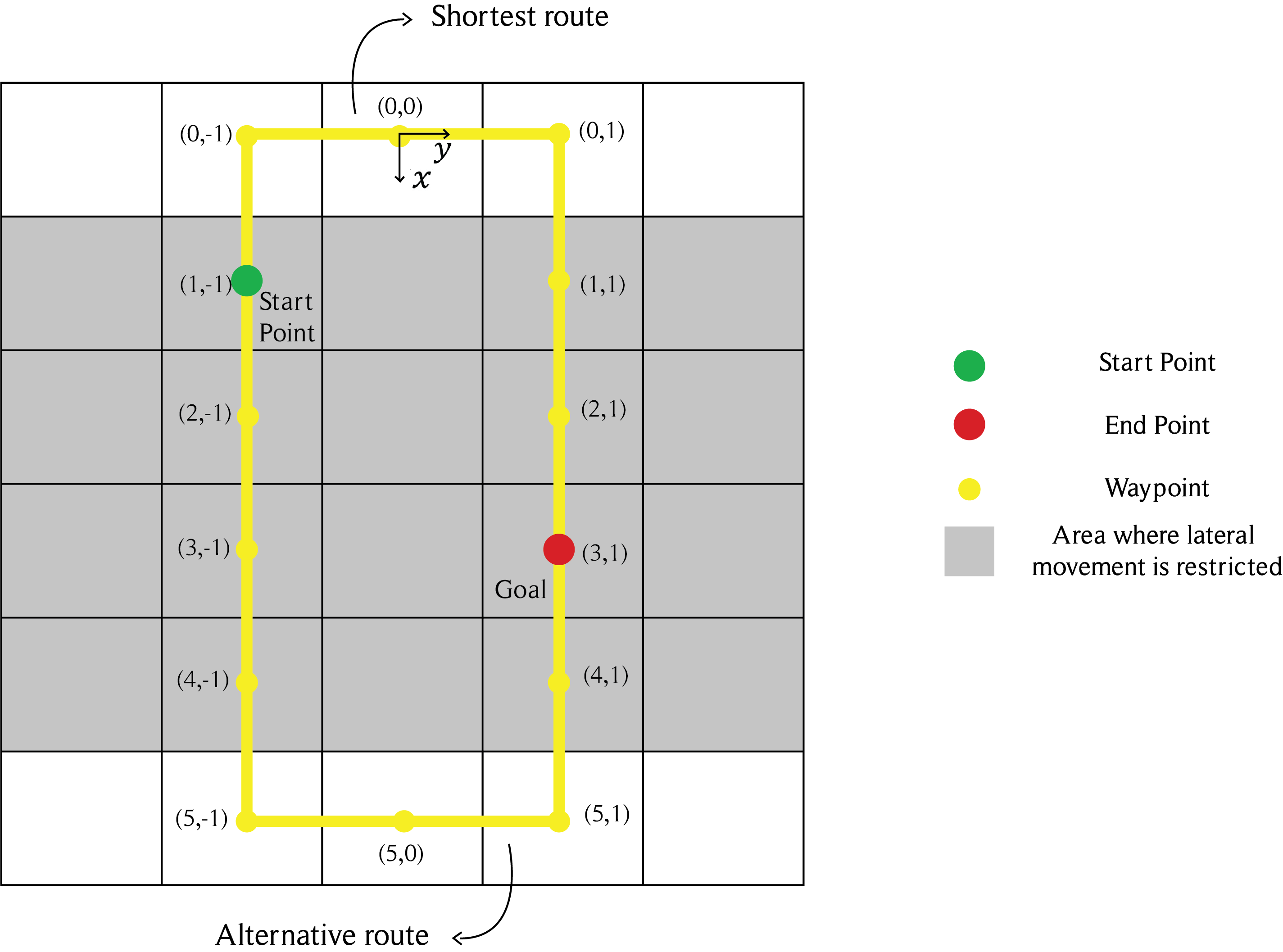}
            }\hspace{1mm}
\caption{Different components in the simulation setup.}
\label{fig:2}
 \end{figure*}

Cameras are added to the front and rear of the robot, which is tilted to focus on the crop rows as shown in Figure \ref{subfig:2_1}. An additional camera has been added to the robot's rear so that it does not need to alter its orientation when entering a new crop row. The crop row detection workflow uses OpenCV's image processing capabilities, as shown in Figure \ref{subfig:2_2}. This method involved cropping the image stream, masking, grayscaling, thresholding, edge detection, and applying the probabilistic Hough line transform to obtain a line across the crop row. This technique for detecting crop rows is independent of crop type since green masking is employed. Crop row track error is calculated between the camera’s centre and this detected line. Crop row track error, as shown in Figure \ref{subfig:2_3}, is simply the absolute difference of $x\;coordinate$ of the camera centre and $x\;coordinate$ of the centre of the line detected. In this paper, we employ a relatively simple method for crop row detection; however, more robust detection methods can be readily incorporated into our navigation strategy for dealing with challenging field conditions. 

\subsection{Training Setup}

\begin{figure}[t]
    \centering
    \includegraphics[scale=0.1]{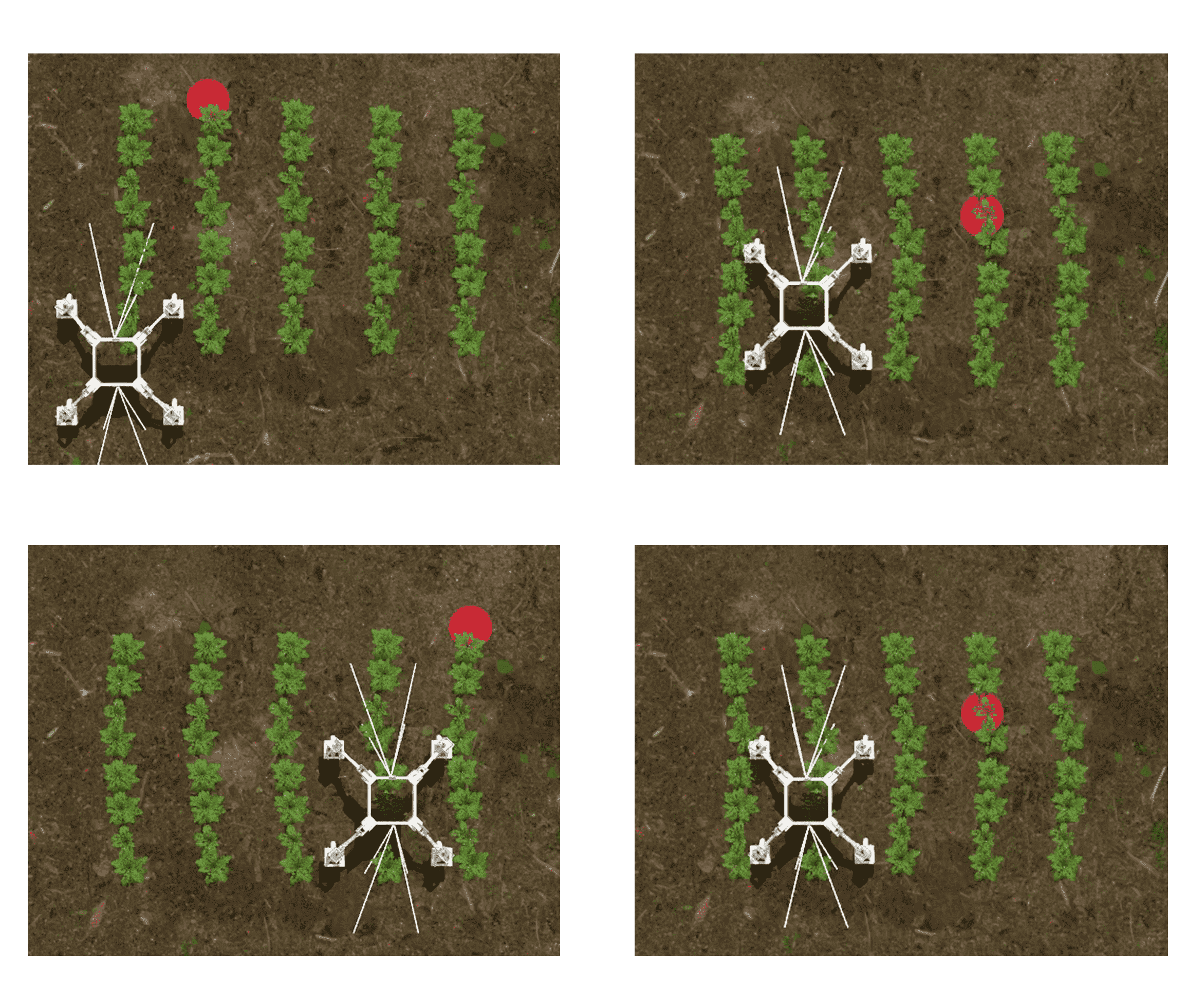}
    \caption{Randomized start and goal locations at the start of each episode. A solid red circle represents the goal location. White guidelines in the robot model represent where the two cameras are directed. }
    \label{fig:3.1}
\end{figure}

Robot and goal positions are randomized at the beginning of each episode. The robot is spawned between crop rows or non-crop areas at random locations, as shown in Figure \ref{fig:3.1}. The goal is produced at random locations within the crop sections. At the start of each episode, crop rows receive slight variations in orientation. These domain randomizations are performed to make the DRL Agent's learning more robust. Each row contains eight plants, and there are five rows in total. To simulate a real-world scenario, the terrain background is chosen to resemble a mud ground. In this training scenario, the crop is considered to be 3D model of potato plants.

The following are the main components of the training setup:

 \subsubsection{Waypoints}

A waypoint is an intermediate reference point that helps us to reach our destination. The whole navigating procedure is simplified with this addition. We create a list of waypoints along the shortest route as given in Figure \ref{subfig:2_4}. Within the crop sections, lateral movement is restricted because it would damage the crops. To promote grid-based movement, we use the Manhattan distance to determine the distance between the robot and the nearest waypoint instead of the Euclidean distance. 

 \subsubsection{Observations}

We provide the position and orientation of the robot, the position of the goal, crop row track error from both cameras, a record of previous actions, and the position and Manhattan distance between the nearest waypoint, along with the values published to each joint of the robot. Before providing the observations to the agent, we normalize it.

\subsubsection{Actions}

In this simulation, we have utilized multiple steering configurations where the choice of steering configuration is dependent on orientation toward the nearest waypoint. If the orientation is horizontal, we select steering for lateral movement. For vertical orientation, we select symmetric four-wheel steering. However, if the orientation error is greater than a certain value, we use zero-turn steering to correct it. For symmetric four-wheel steering, $v_x$ is held constant at $\pm$3.0 m/s, where the sign is changed in accordance with whether the waypoint is ahead or behind the robot. $v_x$ is zero for zero turn and steering for lateral movement, as discussed in the Sec. \ref{robot_setup}. As shown in Table \ref{Table:1}, we only change $\omega$ in a continuous range during each action based on the steering configuration.

\begin{table}[b]
    \caption{Actions}
    \centering
        \begin{tabular}{|c|c|c|c|} \hline 
        
        \textbf{Action} & \textbf{Steering Configuration} &  \textbf{$v_x$} (m/s) & \textbf{$\omega$} (rad/s)  \\ \hline 
        
        Changing $\omega$  &  Symmetric 4WS  & $\pm$3.0  & [-1, +1] \\ \hline  
        Changing $\omega$  &  Steering for lateral movement  & 0 & [-1, +1]*3 \\ \hline 
        Changing $\omega$  &  Zero turn steering  & 0 & [-1, +1]*3 \\ \hline
        
        \end{tabular}
        %\vspace*{5mm}
        \footnotesize{$\omega$ in zero turn and steering for lateral movement is rescaled by a factor of 3}\\
    \label{Table:1}
\end{table}

\subsubsection{Rewards}

\[ r_{dist}\; = - M_d\; from\; nearest\; waypoint\; \]
%\[r_{cam}\;  = - e_{cam1} - e_{cam2} \] 

%\[r_{ctrl}\; = -( \omega \; in\; present\ action  -\omega \; in\; previous\ action) \]
\[r_{ctrl}\; = -\left| \omega_{t-1}-\omega_{t}\right| \]

$$
r_{cam}=\begin{cases}
		+1, & \text{if $e_{cam1}$ or $e_{cam2}$  $\leq$ 40 }\\
            0, & \text{otherwise}\\
		 \end{cases}
$$

$$
r_{wp}=\begin{cases}
			\dfrac{+100}{n_{wp}}, & \text{if $M_{d}$ to nearest waypoint $\leq$ 0.3m }\\
            0, & \text{otherwise}\\
		 \end{cases}
$$

$$
r_{goal}=\begin{cases}
			+100, & \text{if $dist_{goal} \leq$ 0.3m }\\
            -100, & \text{$n_{steps} \geq$ in a episode  500}\\
            -100, & \text{ $M_d \geq$  1.5m}\\
		 \end{cases}
$$

\[ r_{time}\; = -1 \]

$r_{dist}$ is the reward obtained as the robot moves toward the nearest waypoint. $M_d$ is the Manhattan distance. We normalize the above rewards in the range of -1 to +1. $\omega_{t}$, $\omega_{t-1}$ represent the angular velocity predicted by the DRL agent at the present and previous steps. It is to be noted that we have not used the rescaled value of angular velocity for calculating $r_{ctrl}$. This is done to avoid significant variations in subsequent actions so that the robot movement remains smooth. We normalize the above reward in the range of -1 to 0. $e_{cam1}$ and $e_{cam2}$ are the crop row track error from camera one and camera two, respectively. $r_{wp}$ is the reward obtained as the robot reaches a waypoint. $n_{wp}$ is the number of waypoints between the start and endpoint in an episode. $r_{goal}$ is the reward obtained as the robot reaches the goal. $dist_{goal}$ is the distance between the robot's current position and the goal in an episode. $n_{steps}$ indicates the number of steps. Additionally, we give a time penalty of (-1 at each step so that the robot will try to reach the goal in a minimum number of steps.
\[
 r_t\, = r_{dist}\,  \\ +\, r_{cam}\, +\, r_{wp}\, +\, r_{goal}\, +\, r_{ctrl}\, +\, r_{time}\, \\
\]

 Here, $r_t$ represents the total reward an agent gets in a single step. The episode terminates if the Manhattan distance to the nearest waypoint is greater than 1.5m or the distance to the goal is less than 0.3m. An episode is truncated if the number of steps is greater than 500.

 \subsection{Software Setup}
\begin{figure}[b]
    \centering
    \includegraphics[scale=0.35]{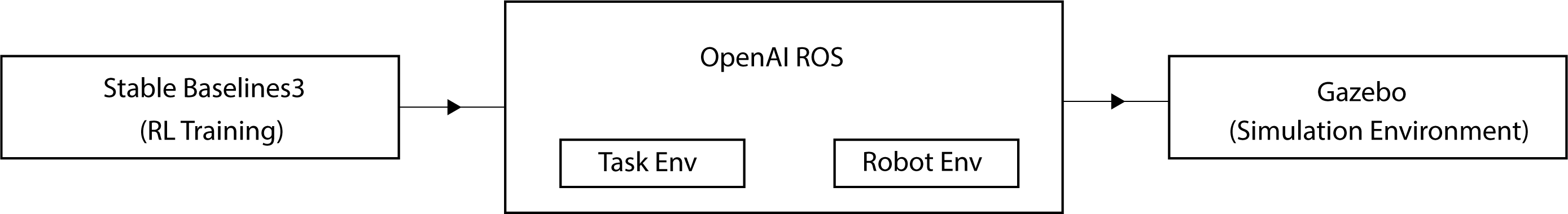}
    \caption{Software Workflow}
    \label{fig:2.9}
\end{figure}

\begin{figure}[b]
    \centering
    \includegraphics[scale=0.06]{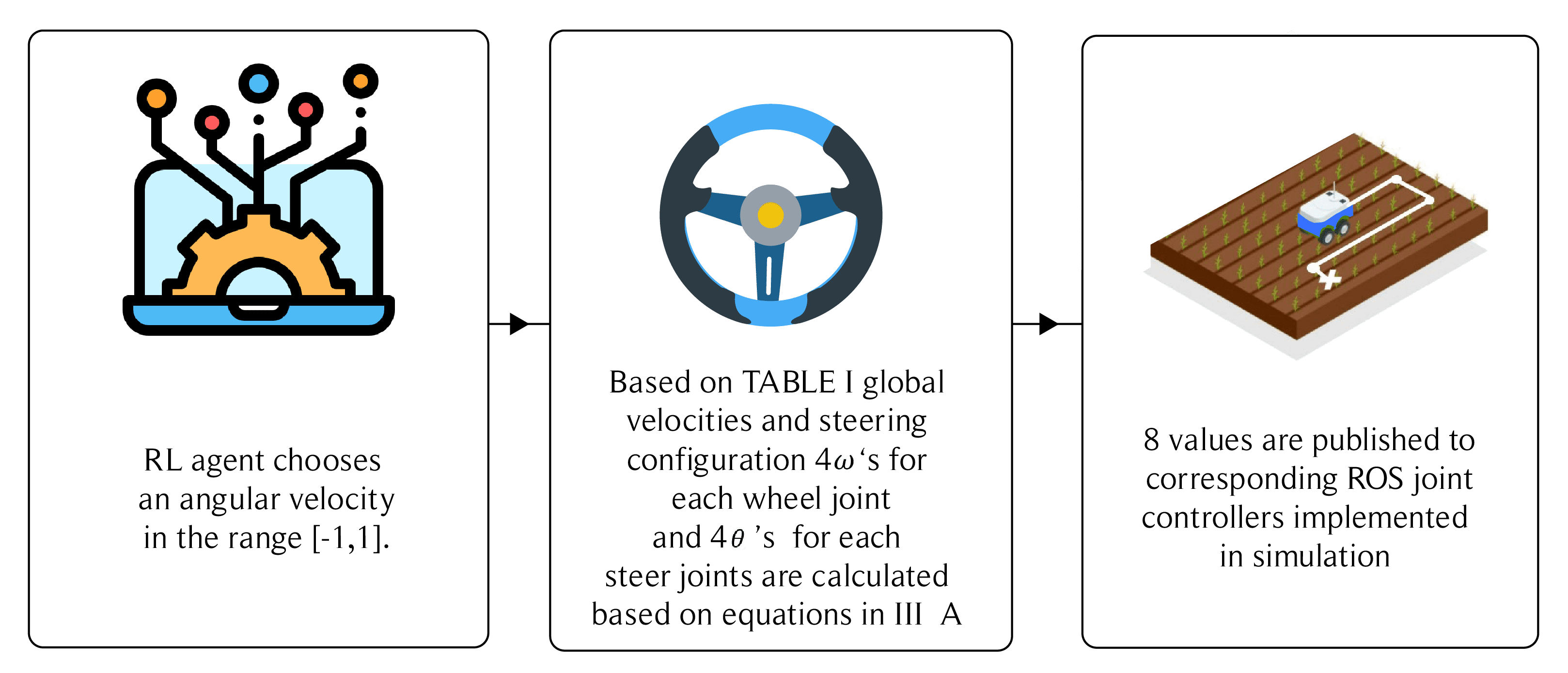}
    \caption{Action Mapping to Simulation.}
    \label{fig:3}
\end{figure}

Simulation Environment setup was carried out in Gazebo\cite{1389727}. We used OpenAI ROS\cite{roswiki} as the DRL framework, which expands Gym's\cite{brockman2016openai} capabilities by allowing seamless interaction with robotic systems in ROS. We used Stable Baselines3\cite{stable-baselines3}, a popular library for implementing state-of-the-art \ac{DRL} algorithms. Figure \ref{fig:2.9} shows how various software layers interact with each other for RL experiments. The training makes use of the four state-of-the-art algorithms SAC (Soft Actor Critic)\cite{haarnoja2018soft}, Twin Delayed DDPG (TD3)\cite{fujimoto2018addressing}, Advantage Actor Critic (A2C)\cite{mnih2016asynchronous} and PPO (Proximal Policy Optimization)\cite{schulman2017proximal} algorithm from stable baselines3. Figure \ref{fig:3} describes how the change will be reflected in the robot if the agent selects an action.

%\begin{comment}Transition Dynamics shown in
%\end{comment}

\section{SIMULATIONS AND RESULTS}

\subsection{Crop Row Tracking using DRL}

We ran an initial test with symmetric \ac{4WS} steering and constant forward linear velocity to determine if crop row tracking is feasible. The purpose of the simulation was to train the robot to reach the goal while precisely following an irregularly curved crop row by varying its angular velocity. The robot's initial distance from the target was 9 meters. There was just a single camera in this simulation. If the crop row track error was less than 40, we gave a reward of +1 and a punishment of -1 time for each time step. +10 was also given as a reward if the robot reached its goal.

\begin{figure}[t]
    \centering
    %\includesvg[scale=0.2]{single_row_combined.svg}
    \includegraphics[width=0.3\textwidth]{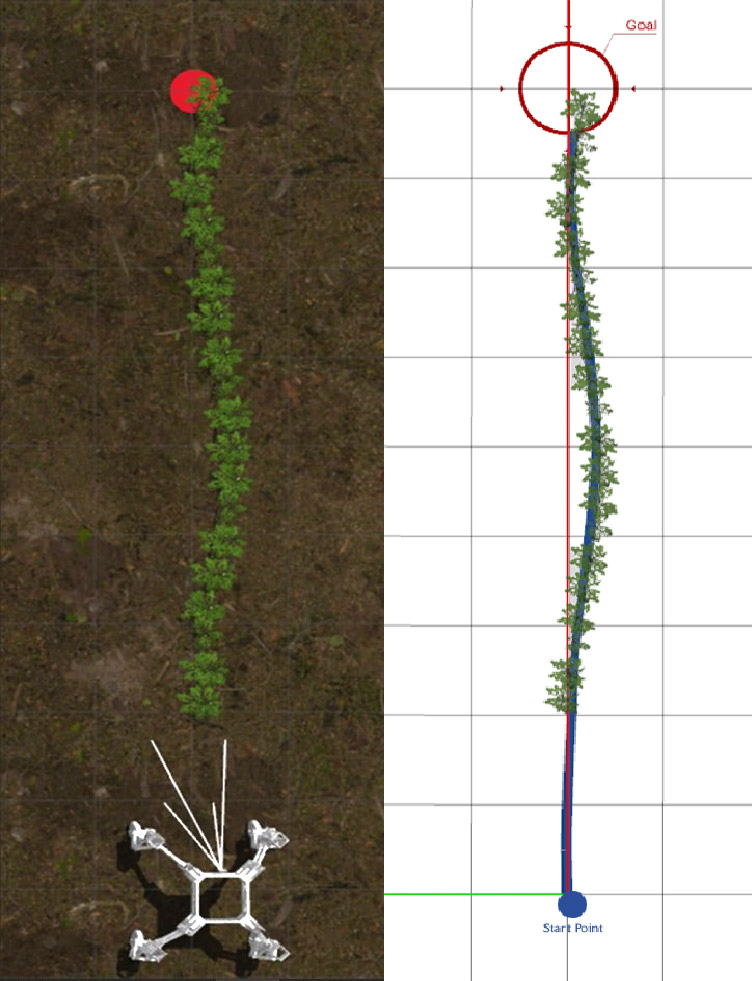}
    \caption{Results of crop row tracking with a single crop row. The blue line is the path followed by the centre of the robot. Green Patches are crops located in the simulation setup. The start point and goal location are also marked.}
    \label{fig:4}
\end{figure}

\begin{figure}[t]
    \centering
    \includegraphics[scale=0.45]{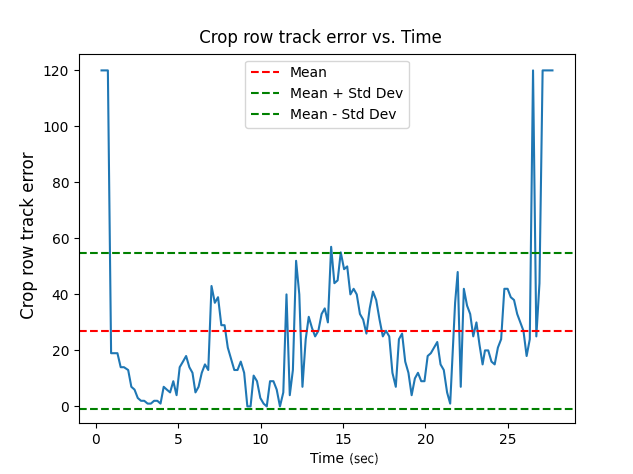}
    \caption{Crop row track error values over navigation towards goal in a single iteration.}
    \label{fig:5}
\end{figure}

 The blue line in Figure \ref{fig:4} is the path followed by the centre of the robot. Green Patches are crops located in the simulation setup. From Figure \ref{fig:4}, it is clear that the robot was able to track the crop row precisely. Figure \ref{fig:5} shows the entire crop row track error that occurred while the robot navigated toward the goal in a single iteration. To promote the detection of crop rows, we set the default crop row error value to 120, which is half the width of the cropped image stream from the camera. So, as soon as a crop row is detected, this error value goes down. The graph in Figure \ref{fig:5} shows considerable variations in crop row error at the start and the end because the robot enters and leaves the crop row. Some of the significant takeaways from the simulations over 500 iterations are given in Table \ref{Table:2}.

   \begin{table}[b]
    \centering
        \caption{Results over 500 iterations}
        \label{Table:5_7}
        \begin{tabular}{|l|l|} \hline 
        
         Path length & 8.71 m   \\ \hline 
         Time Taken  & 28.0 sec \\ \hline 
         Mean Crop row track error & 33.15  \\ \hline 
         Standard Deviation of Crop row track error & $\pm$31.39  \\ \hline
        
        \end{tabular}
    \label{Table:2}
\end{table}

\subsection{Multiple Crop Row Navigation}

\begin{figure}[b!]
    \centering
    \includegraphics[scale=0.4]{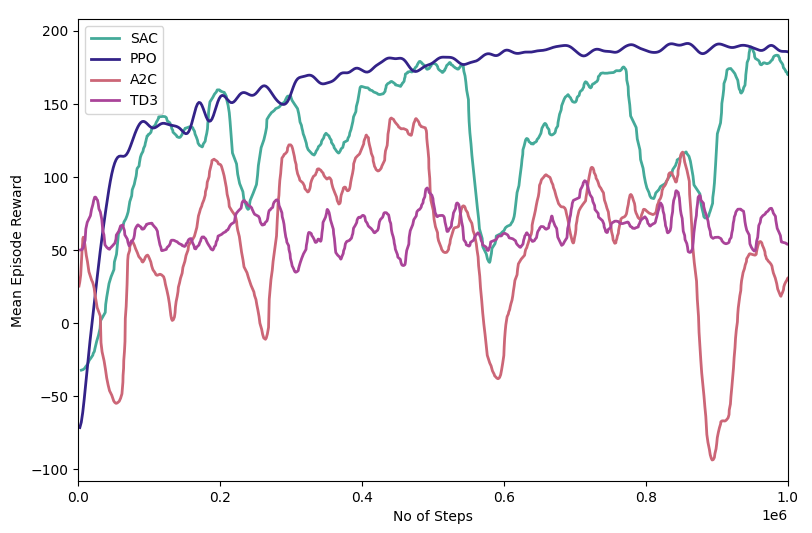}
    \caption{Mean Episode Reward over 1 million steps.}
    \label{fig:7}
\end{figure}

\begin{figure}[b!]
    \centering
    \includegraphics[scale=0.36]{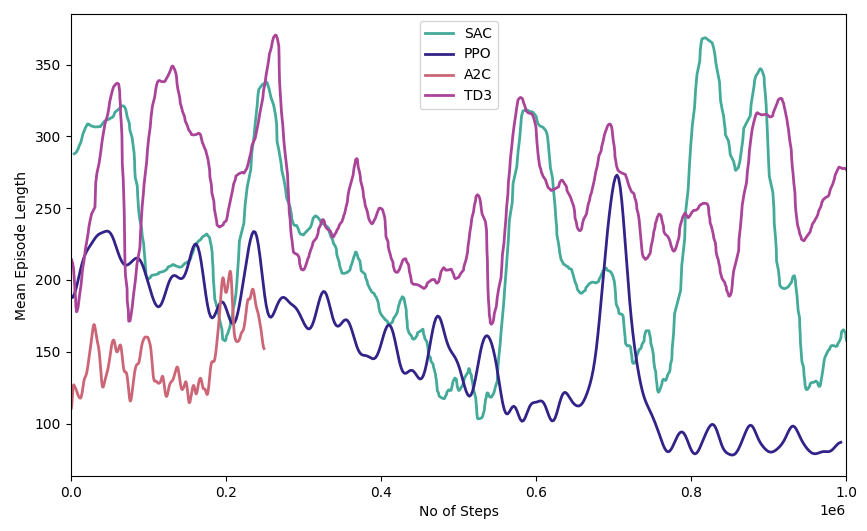}
    \caption{Mean Episode Length over 1 million steps.}
    \label{fig:8}
\end{figure}

 As depicted in Figure \ref{fig:7}, the training was carried out for all four algorithms for more than 1 million steps. The best performance was achieved with PPO. SAC, A2C, and TD3 all provided inconsistent rewards and extended episodes throughout the duration of the training. A2C and TD3 had the worst performance. In Figure \ref{fig:8}, we see that the mean episode length was shortest when PPO was used. The average episode lasted below 100 steps, and the average reward remained above 180 over 1 million steps. Hyperparameters of PPO used from the stable baseline are shown in Table \ref{tab:ppo_hyperparameters}.
 
 \begin{table}[b!]
    \centering
    \caption{PPO Hyperparameters}
    \label{tab:ppo_hyperparameters}
    \begin{tabular}{|c|c|} \hline 
        
        \textbf{Hyperparameter} & \textbf{Value} \\ \hline 
        
        Learning Rate & 0.0003 \\ \hline 
        Number of Steps & 2048 \\ \hline 
        Batch Size & 64 \\ \hline 
        Number of Epochs & 10 \\ \hline 
        Gamma & 0.99 \\ \hline 
        GAE Lambda & 0.95 \\ \hline 
        Clip Range & 0.2 \\ \hline
        
    \end{tabular}
    
\end{table} 
To analyze the result, we gathered 420 data points that register the time and distance traveled by the robot to reach the goal. In these 420 cases, the goal and robot's starting position are randomly set. The obtained results are presented in Table \ref{Table:3}.
\vspace{0mm}
 \begin{figure}[t!]
    \centering
    \includegraphics[scale=0.2]{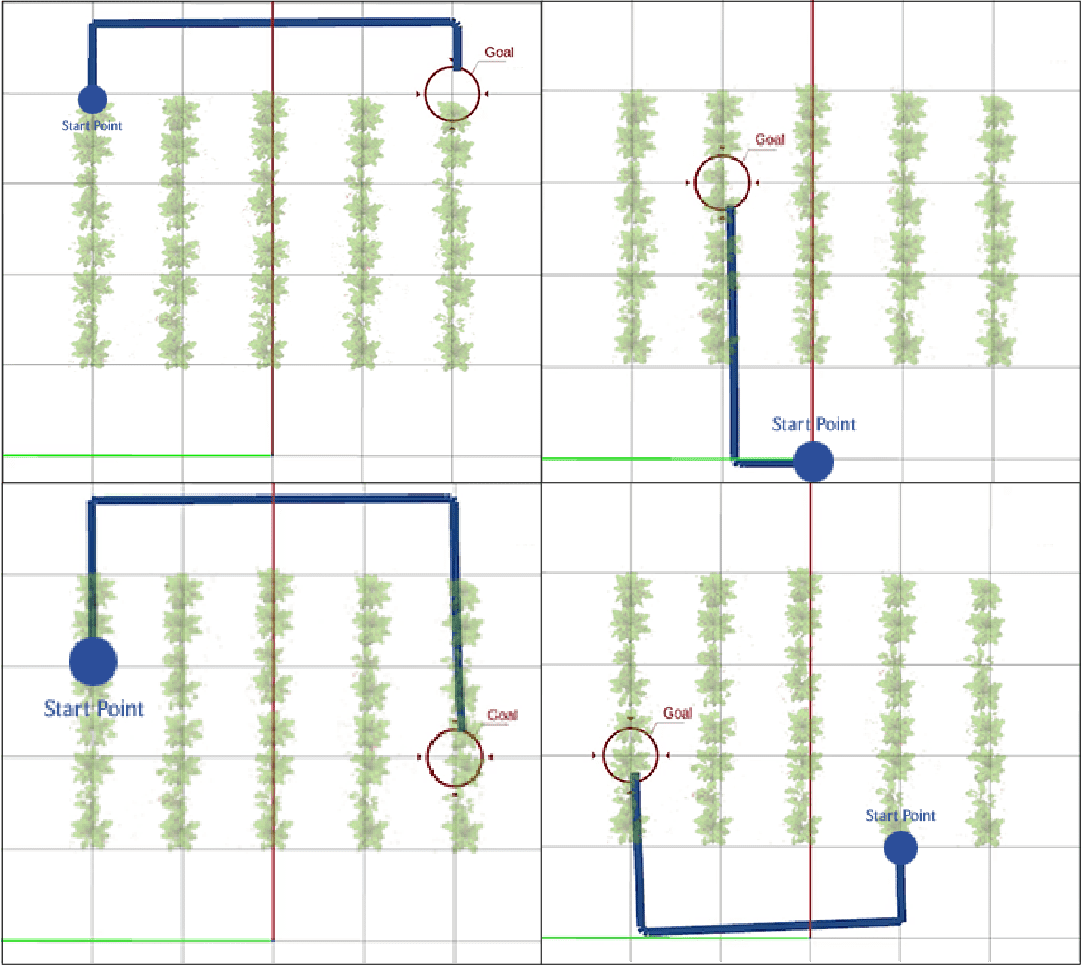}
    \caption{Results of Navigation in the field with multiple crop rows. The blue line is the path followed by the center of the robot. Green Patches are crops located in the simulation setup. The start point and goal location are also marked.}
    \label{fig:6}
\end{figure}

\begin{table}[h!]
    \centering
        \caption{Results}
        \label{Table:5_10}
        \begin{tabular}{|l|l|} \hline 
         Success Percentage & 83.57\% \\ \hline 
         Mean Distance Travelled & 4.59 m \\ \hline 
         Mean Manhattan Distance & 4.93 m \\ \hline 
         Mean Time Taken  & 17.3 sec \\ \hline
        \end{tabular}
    \label{Table:3}
\end{table}

There were 351 successful trials out of a total of 420. The mean path length and time were calculated based on the successful trials. It was found to be 4.59m and 17.3s, respectively. Mean Manhattan distance was also computed, which was the distance the robot assigned to travel. The mean Manhattan distance was 4.93m, larger than the mean distance actually travelled. Consequently, our DRL-based autonomous navigation strategy reduced the distance it was assigned to travel. This suggests that our approach is not only efficient but also conservative in its estimations, which could lead to potential resource savings and improved overall performance.

\begin{figure}[ht]
    \centering
    \includegraphics[scale=0.08]{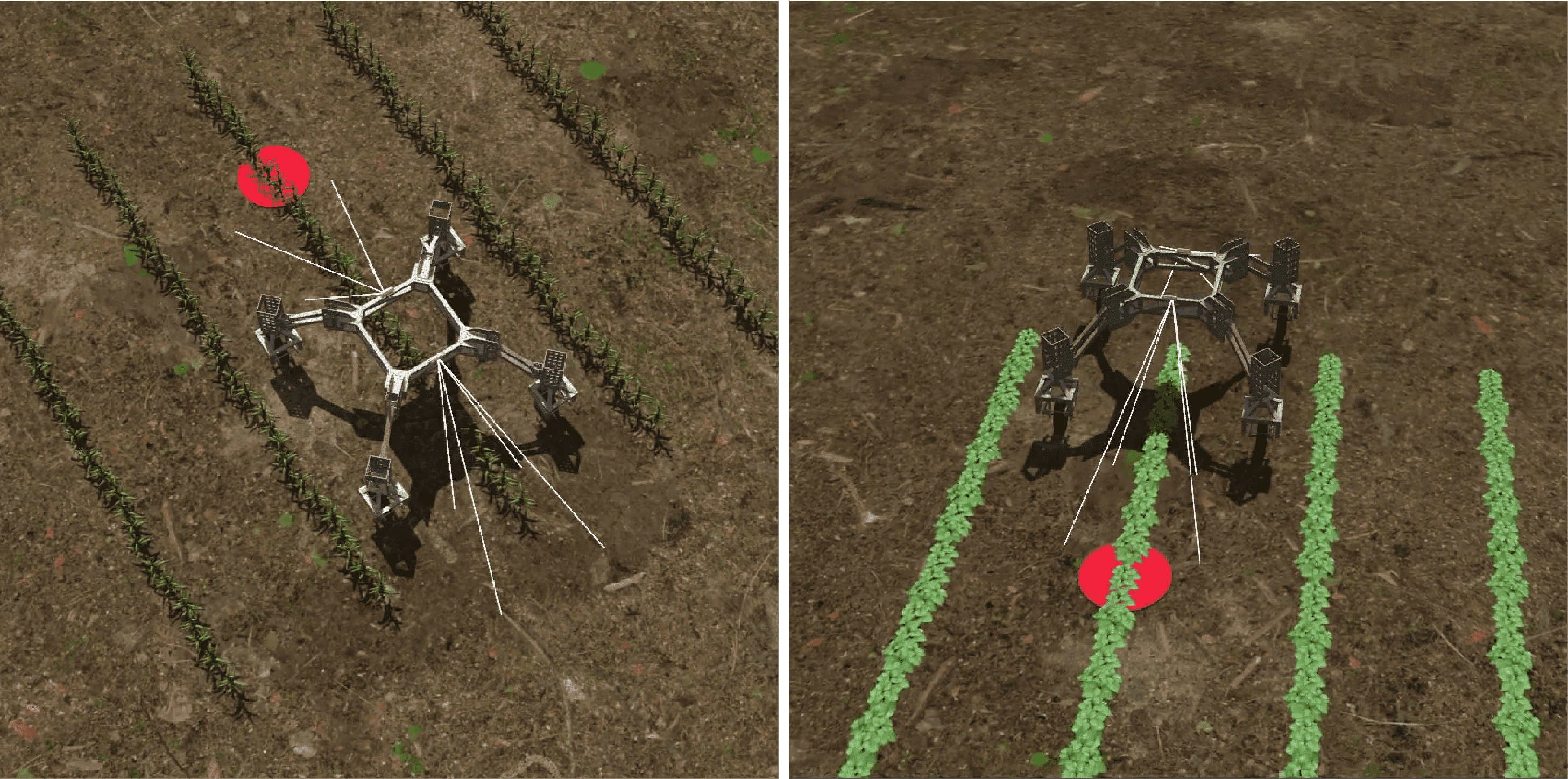}
    \caption{Robot navigating in simulation environment of maize and basil plants.}
    \label{fig:11}
\end{figure}

From Figure \ref{fig:6}, it is clear that the robot was able to track the crop row precisely and reach the respective destinations. Moreover, the successful testing of the trained model with slightly increased field sizes and previously unseen crops such as basil and maize, depicted in Figure \ref{fig:11}, highlights the adaptability and generic nature of our approach. This versatility is crucial for real-world deployment scenarios where environmental conditions may vary.

\subsection{Comparative analysis with PD Controller on a skid-steered robot}
This section compares the performance of a 4WIS4WID robot with an RL navigation controller to a skid steer configuration robot with a PD controller for linear and angular movement while navigating a C-shaped path having a $2m$ edge length. We used the same robot model for skid steer configuration. The implementation of this was done with the help of the skid steer dive plugin from the gazebo.
 \begin{figure}[t!]
        \centering
            \subfigure[Skid steer PD controller]
            {
                \label{subfig:521_1}
                \includegraphics[scale=0.5]{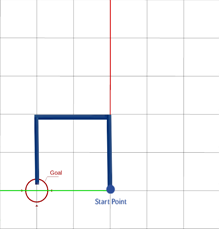} 
                
            } 
            \subfigure[4WIS4WID RL Navigation strategy]
            {
                \label{subfig:521_2}
                \includegraphics[scale=0.5]{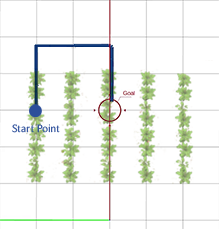}
            }
\caption{Comparison between RL and PD controller in 4WIS4WID and skid steer.}
\label{figch5_21}
 \end{figure}
\begin{table}[b]
    \centering
    \caption{Results of Comparison}
    \label{Table:ch5_19}
    \begin{tabular}{|c|c|c|} \hline 
        \textbf{Type of Navigation} & \textbf{Distance Travelled(m)} & \textbf{Time Taken(s)} \\ \hline 
        RL Navigation(4WIS4WID) & 5.29 & 18.0 \\ \hline 
        PD Controller(Skid Steer) & 5.97 & 49.4 \\ \hline
    \end{tabular}
\end{table}

From Table \ref{Table:ch5_19}, it is clear that in the case of the skid steer configuration with the PD controller, the robot required more time to complete the C-shaped path because it had to rotate first in order to move forward. In contrast, the 4WIS4WID configuration has the advantage of additional lateral steering, which requires the robot to move immediately toward the next waypoint. The robot travels a shorter distance (0.68m) than the total grid-based distance assigned to travel, as confirmed by the fact that the distance tracked by the RL-based navigation strategy is shorter than that of the PD-controlled skid steer robot.

\section{CONCLUSIONS}
 This paper's pursuit of integrating crop row detection and autonomous navigation into a single system is significant because it represents a robust combination of the two fields of study. This navigation strategy does not require a separate controller for each steering configuration or distinct tasks, such as switching crop rows. This end-to-end system uses DRL's decision-making ability to predict an angular velocity that efficiently navigates the 4WIS4WID mobile robot between multiple crop rows to a goal without considering path planning and path tracking to be separate processes. With zero-turn steering, we were able to adjust the orientation of our robot, and by utilizing lateral steering, we could easily transition between crop rows. Symmetric \ac{4WS} steering enabled us to take tight turns with precision. Furthermore, the equations derived in this paper also offer a robust framework for achieving teleoperation across various 4WIS4WID steering configurations using either a keyboard or joystick interface.

Everything in this paper was done in a simulation environment. In future, we will implement this navigation strategy on a physical 4WIS4WID mobile robot capable of traversing real-world terrain. By pursuing this sim-real transfer, we aim to explore the feasibility and effectiveness of our navigation strategy in real environments, with all the complexities and uncertainties they entail.  

\section*{Acknowledgment}
This research was supported by the \acf{DST}-\acf{TDP}, Government of India, under project number DST/ME/2020009.

\bibliographystyle{IEEEtran}
\bibliography{IEEEabrv,Bibilography}

% Generated by IEEEtran.bst, version: 1.14 (2015/08/26)
\begin{thebibliography}{10}
\providecommand{\url}[1]{#1}
\csname url@samestyle\endcsname
\providecommand{\newblock}{\relax}
\providecommand{\bibinfo}[2]{#2}
\providecommand{\BIBentrySTDinterwordspacing}{\spaceskip=0pt\relax}
\providecommand{\BIBentryALTinterwordstretchfactor}{4}
\providecommand{\BIBentryALTinterwordspacing}{\spaceskip=\fontdimen2\font plus
\BIBentryALTinterwordstretchfactor\fontdimen3\font minus \fontdimen4\font\relax}
\providecommand{\BIBforeignlanguage}[2]{{%
\expandafter\ifx\csname l@#1\endcsname\relax
\typeout{** WARNING: IEEEtran.bst: No hyphenation pattern has been}%
\typeout{** loaded for the language `#1'. Using the pattern for}%
\typeout{** the default language instead.}%
\else
\language=\csname l@#1\endcsname
\fi
#2}}
\providecommand{\BIBdecl}{\relax}
\BIBdecl

\bibitem{RePEc:ags:faoeff:319842}
\BIBentryALTinterwordspacing
Food and A.~O. of~the United Nations~(FAO), ``{The future of food and agriculture – Alternative pathways to 2050},'' Food and Agriculture Organization of the United Nations, Agricultural Development Economics Division (ESA), The Future of Food and Agriculture 319842, undated. [Online]. Available: \url{https://ideas.repec.org/p/ags/faoeff/319842.html}
\BIBentrySTDinterwordspacing

\bibitem{https://doi.org/10.1049/joe.2014.0241}
\BIBentryALTinterwordspacing
M.-H. Lee and T.-H.~S. Li, ``Kinematics, dynamics and control design of 4wis4wid mobile robots,'' \emph{The Journal of Engineering}, vol. 2015, no.~1, pp. 6--16, 2015. [Online]. Available: \url{https://ietresearch.onlinelibrary.wiley.com/doi/abs/10.1049/joe.2014.0241}
\BIBentrySTDinterwordspacing

\bibitem{BAI2023107584}
\BIBentryALTinterwordspacing
Y.~Bai, B.~Zhang, N.~Xu, J.~Zhou, J.~Shi, and Z.~Diao, ``Vision-based navigation and guidance for agricultural autonomous vehicles and robots: A review,'' \emph{Computers and Electronics in Agriculture}, vol. 205, p. 107584, 2023. [Online]. Available: \url{https://www.sciencedirect.com/science/article/pii/S0168169922008924}
\BIBentrySTDinterwordspacing

\bibitem{ZINEELABIDINE202130}
\BIBentryALTinterwordspacing
M.~Zine-El-Abidine, H.~Dutagaci, G.~Galopin, and D.~Rousseau, ``Assigning apples to individual trees in dense orchards using 3d colour point clouds,'' \emph{Biosystems Engineering}, vol. 209, pp. 30--52, 2021. [Online]. Available: \url{https://www.sciencedirect.com/science/article/pii/S1537511021001392}
\BIBentrySTDinterwordspacing

\bibitem{MATAS2000119}
\BIBentryALTinterwordspacing
J.~Matas, C.~Galambos, and J.~Kittler, ``Robust detection of lines using the progressive probabilistic hough transform,'' \emph{Computer Vision and Image Understanding}, vol.~78, no.~1, pp. 119--137, 2000. [Online]. Available: \url{https://www.sciencedirect.com/science/article/pii/S1077314299908317}
\BIBentrySTDinterwordspacing

\bibitem{9096177}
L.~C. Santos, F.~N. Santos, E.~J. Solteiro~Pires, A.~Valente, P.~Costa, and S.~Magalhães, ``Path planning for ground robots in agriculture: a short review,'' in \emph{2020 IEEE International Conference on Autonomous Robot Systems and Competitions (ICARSC)}, 2020, pp. 61--66.

\bibitem{8228012}
M.~A. Juman, Y.~W. Wong, R.~K. Rajkumar, and C.~Y. H'ng, ``An integrated path planning system for a robot designed for oil palm plantations,'' in \emph{TENCON 2017 - 2017 IEEE Region 10 Conference}, 2017, pp. 1048--1053.

\bibitem{santos_santos_mendes_costa_lima_reis_shinde_2020}
L.~Santos, F.~Santos, J.~Mendes, P.~Costa, J.~Lima, R.~Reis, and P.~Shinde, ``Path planning aware of robot’s center of mass for steep slope vineyards,'' \emph{Robotica}, vol.~38, no.~4, p. 684–698, 2020.

\bibitem{8456505}
X.~Gao, J.~Li, L.~Fan, Q.~Zhou, K.~Yin, J.~Wang, C.~Song, L.~Huang, and Z.~Wang, ``Review of wheeled mobile robots’ navigation problems and application prospects in agriculture,'' \emph{IEEE Access}, vol.~6, pp. 49\,248--49\,268, 2018.

\bibitem{PATLE2019582}
\BIBentryALTinterwordspacing
B.~Patle, G.~{Babu L}, A.~Pandey, D.~Parhi, and A.~Jagadeesh, ``A review: On path planning strategies for navigation of mobile robot,'' \emph{Defence Technology}, vol.~15, no.~4, pp. 582--606, 2019. [Online]. Available: \url{https://www.sciencedirect.com/science/article/pii/S2214914718305130}
\BIBentrySTDinterwordspacing

\bibitem{JIANG202017}
\BIBentryALTinterwordspacing
Z.~Jiang, J.~Zhu, Z.~Lin, Z.~Li, and R.~Guo, ``3d mapping of outdoor environments by scan matching and motion averaging,'' \emph{Neurocomputing}, vol. 372, pp. 17--32, 2020. [Online]. Available: \url{https://www.sciencedirect.com/science/article/pii/S0925231219312846}
\BIBentrySTDinterwordspacing

\bibitem{10.1109/ROBIO49542.2019.8961753}
\BIBentryALTinterwordspacing
P.~Gao, Z.~Liu, Z.~Wu, and D.~Wang, ``A global path planning algorithm for robots using reinforcement learning,'' in \emph{2019 IEEE International Conference on Robotics and Biomimetics (ROBIO)}.\hskip 1em plus 0.5em minus 0.4em\relax IEEE Press, 2019, p. 1693–1698. [Online]. Available: \url{https://doi.org/10.1109/ROBIO49542.2019.8961753}
\BIBentrySTDinterwordspacing

\bibitem{9419029}
H.~Sun, W.~Zhang, R.~Yu, and Y.~Zhang, ``Motion planning for mobile robots—focusing on deep reinforcement learning: A systematic review,'' \emph{IEEE Access}, vol.~9, pp. 69\,061--69\,081, 2021.

\bibitem{8468643}
K.~Zhang, F.~Niroui, M.~Ficocelli, and G.~Nejat, ``Robot navigation of environments with unknown rough terrain using deep reinforcement learning,'' in \emph{2018 IEEE International Symposium on Safety, Security, and Rescue Robotics (SSRR)}, 2018, pp. 1--7.

\bibitem{9468918}
H.~Hu, K.~Zhang, A.~H. Tan, M.~Ruan, C.~Agia, and G.~Nejat, ``A sim-to-real pipeline for deep reinforcement learning for autonomous robot navigation in cluttered rough terrain,'' \emph{IEEE Robotics and Automation Letters}, vol.~6, no.~4, pp. 6569--6576, 2021.

\bibitem{9197114}
A.~Ahmadi, L.~Nardi, N.~Chebrolu, and C.~Stachniss, ``Visual servoing-based navigation for monitoring row-crop fields,'' in \emph{2020 IEEE International Conference on Robotics and Automation (ICRA)}, 2020, pp. 4920--4926.

\bibitem{Jazar2008}
\BIBentryALTinterwordspacing
R.~N. Jazar, \emph{Steering Dynamics}.\hskip 1em plus 0.5em minus 0.4em\relax Boston, MA: Springer US, 2008, pp. 379--454. [Online]. Available: \url{https://doi.org/10.1007/978-0-387-74244-1\_7}
\BIBentrySTDinterwordspacing

\bibitem{1389727}
N.~Koenig and A.~Howard, ``Design and use paradigms for gazebo, an open-source multi-robot simulator,'' in \emph{2004 IEEE/RSJ International Conference on Intelligent Robots and Systems (IROS) (IEEE Cat. No.04CH37566)}, vol.~3, 2004, pp. 2149--2154 vol.3.

\bibitem{roswiki}
\BIBentryALTinterwordspacing
``openai ros ros wiki.'' [Online]. Available: \url{http://wiki.ros.org/openai\_ros}
\BIBentrySTDinterwordspacing

\bibitem{brockman2016openai}
G.~Brockman, V.~Cheung, L.~Pettersson, J.~Schneider, J.~Schulman, J.~Tang, and W.~Zaremba, ``Openai gym,'' 2016.

\bibitem{stable-baselines3}
\BIBentryALTinterwordspacing
A.~Raffin, A.~Hill, A.~Gleave, A.~Kanervisto, M.~Ernestus, and N.~Dormann, ``Stable-baselines3: Reliable reinforcement learning implementations,'' \emph{Journal of Machine Learning Research}, vol.~22, no. 268, pp. 1--8, 2021. [Online]. Available: \url{http://jmlr.org/papers/v22/20-1364.html}
\BIBentrySTDinterwordspacing

\bibitem{haarnoja2018soft}
T.~Haarnoja, A.~Zhou, P.~Abbeel, and S.~Levine, ``Soft actor-critic: Off-policy maximum entropy deep reinforcement learning with a stochastic actor,'' 2018.

\bibitem{fujimoto2018addressing}
S.~Fujimoto, H.~van Hoof, and D.~Meger, ``Addressing function approximation error in actor-critic methods,'' 2018.

\bibitem{mnih2016asynchronous}
V.~Mnih, A.~P. Badia, M.~Mirza, A.~Graves, T.~P. Lillicrap, T.~Harley, D.~Silver, and K.~Kavukcuoglu, ``Asynchronous methods for deep reinforcement learning,'' 2016.

\bibitem{schulman2017proximal}
J.~Schulman, F.~Wolski, P.~Dhariwal, A.~Radford, and O.~Klimov, ``Proximal policy optimization algorithms,'' 2017.

\end{thebibliography}
\end{document}